\documentclass[11pt]{article}
\usepackage{eamt23}
\usepackage{times}
\usepackage{url}
\usepackage{CJKutf8}
\usepackage{latexsym}
\usepackage{float}
\usepackage{amsmath}
\usepackage{titlesec}
\usepackage[small,bf]{caption} 
\usepackage{dblfloatfix}
\titleformat{\paragraph}
{\normalfont\normalsize\bfseries}{\theparagraph}{1em}{}
\titlespacing*{\paragraph}
{0pt}{3.25ex plus 1ex minus .2ex}{1.5ex plus .2ex}

\title{Evaluation of Chinese-English Machine Translation of Emotion-Loaded Microblog Texts: A Human Annotated Dataset for the Quality Assessment of Emotion Translation}

\author{Shenbin Qian$^1$, \ Constantin Orăsan$^1$, \ Félix do Carmo$^1$,\\
  \textbf{Qiuliang Li}$^2$, \textbf{Diptesh Kanojia}$^3$\\
  Centre for Translation Studies, University of Surrey, UK$^1$ \\
  Xi'an International Studies University, China$^2$ \\
  Department of Computer Science, University of Surrey, UK$^3$ \\
  \texttt{\{s.qian, c.orasan, f.docarmo, d.kanojia\}@surrey.ac.uk$^{1,3}$} \\ 
  qiuliang0909@gmail.com$^2$}

\date{}

\begin{document}
\begin{CJK}{UTF8}{gbsn}
\maketitle
\begin{abstract}
In this paper, we focus on how current Machine Translation (MT) tools perform on the translation of emotion-loaded texts by evaluating outputs from Google Translate according to a framework proposed in this paper. We propose this evaluation framework based on the Multidimensional Quality Metrics (MQM) and perform a detailed error analysis of the MT outputs. From our analysis, we observe that about 50\% of the MT outputs fail to preserve the original emotion. After further analysis of the errors, we find that emotion carrying words and linguistic phenomena such as polysemous words, negation, abbreviation \textit{etc.}, are common causes for these translation errors.
\end{abstract}

\section{Introduction} \label{sec1}

To express feelings and attitudes is one of language's major functions~\cite{Waugh1980}. In this digital age, people can easily share their emotions or opinions online using social media platforms. This results in the generation of a large amount of emotion-loaded and opinionated texts. It is important to convey the correct emotion or opinion in the text to a large audience from different linguistic or cultural backgrounds for cross-cultural communication. Otherwise, misinformation or even toxic emotions~\cite{Frost2003} can permeate cross-cultural communication, resulting in harmful implications for the parties involved. Due to the asynchronous nature and sheer quantity of this generated text online, it is impossible for human translators to be present in the loop and perform accurate translations. Hence, machine translation (MT) remains the only viable choice for the task of translating emotion-loaded microblog texts~\cite{Carrera2009}. 

Social media texts on Sina Weibo\footnote{\url{https://weibo.com/}}, the Chinese microblog platform, have their unique characteristics due to certain features of the Chinese language. Since Chinese is a tonal language, there are many characters which share the exact same or very similar pronunciation but with drastically different meanings. Chinese netizens commonly use this language phenomenon to create emotional slang by replacing the original character/word with a homophone character/word to avoid censorship. Similarly, substitution with homographs is another way to create slang, as Chinese is a hieroglyphic language. For example, using ``目田'', which means ``eye field'', and substituting them for ``自由'', meaning ``freedom'' is an example of homograph substitution~\cite{King2013}. We can observe that ``目田'' looks very similar to ``自由'', where a few strokes of the two characters are omitted to refer to the lack of freedom. Abbreviation of long expressions or transliteration of Chinese characters is another observed phenomenon in social media texts. Such features in this new online language variant pose severe challenges to the MT of Chinese social media texts, especially the emotion-loaded and opinionated microblogs. These challenges are different from the ones observed in translating tweets with hashtags or non-standard orthography present in the other languages~\cite{Saadany2021a}. 

There are several studies and datasets which focus on the translation of social media texts, such as TweetMT~\cite{SanVicente2016}, the tweet corpus proposed by Mubarak et al.~\shortcite{Mubarak2020} and the Weibo corpus developed by Ling et al.~\shortcite{Ling2013}. However, none of these focus on the translation of emotions. To the best of our knowledge, there is no research which focuses on the Chinese-English machine translation (C-E MT) of emotion-loaded texts. We endeavour to make our contributions to this area as summarised below:
\begin{itemize}
\itemsep 0mm
\item A quality assessment framework for the machine translation of emotion-loaded texts is proposed for evaluating the MT quality in terms of emotion preservation.
\item A detailed error analysis is performed to find out linguistic phenomena that are more likely to cause C-E MT errors in terms of emotions.
\item A dataset\footnote{https://github.com/shenbinqian/HADQAET}, annotated with translation errors and severity levels, is released to support tasks like error detection and quality estimation of emotion translation.
\end{itemize}

Section~\ref{sec2} describes the related literature in emotion translation and quality assessment of MT. Our proposed framework for human evaluation of the MT quality of emotion-loaded texts is described in Section~\ref{sec3}. In Section~\ref{sec4}, we introduce the dataset and methodology for quality assessment. The result of human evaluation and error analysis is presented and analysed in Section~\ref{sec5}. Section~\ref{sec6} discusses the conclusion and future plan after summarising the whole paper. 

\section{Related Work} \label{sec2}

\subsection{Translation of Emotions and Emotion-Loaded Texts}

The awareness of emotions in translation has been discussed in the early stages of translation studies when the emotional reaction of the reader was of significance in the translation of the Bible~\cite{Lehr2020}. Nida and Taber~\shortcite{Nida1969} emphasised the importance of transferring emotional elements from source to target and proposed to translate the emotionality of the text with a focus on the final translation product. 

Many studies focused on the emotional difference or emotion translation between languages, most of which emphasised on the translation of emotion lexica. Russell and Sato~\shortcite{Russell1995} compared 14 emotional words such as `happy' or `sad' in English, Chinese and Japanese to observe similarities and differences post-translation. Choi and Han~\shortcite{Choi2008} raised concerns about cross-cultural communication regarding the difficulty of finding the equivalence of some emotional concepts such as \textit{`shimcheong'} (a combination of empathy, sympathy, and compassion) in Korean. Similarly, Hurtado de Mendoza et al.~\shortcite{Hurtado2010} also raised questions about one-to-one translations of emotion concepts like `shame' in English and Spanish. For other language pairs like English and Arabic, Kayyal and Russell~\shortcite{Kayyal2013} did very similar studies and found that only one pair (happiness-farah) passed their equivalence tests, and other lexical pairs differed in terms of culture and language. For English and Indonesian, the emotion `happy' can be translated into several different words including `\textit{bahagia}', `\textit{senang}', `\textit{suka}', `\textit{lega}', `\textit{kesenangan}', `\textit{gembira ria}', `\textit{riang}', `\textit{ceria}', `\textit{patah hati}', and `\textit{tenteram}'~\cite{Suryasa2019}. They are not the same in meaning or style, so translating such words might lead to subtle emotional differences in the target language. 

These studies reveal the challenges and importance of translating emotions or emotional lexica in cross-cultural communication. But very few studies focused on machine translation or the quality of machine translation regarding emotion preservation. Mohammad et al.~\shortcite{Mohammad2016} examined sentiments in social media posts in both Arabic-English and English-Arabic translations, and they found that the change of sentiment was mainly caused by ambiguous words, sarcasm, metaphors, and word-reordering issues. Shalunts et al.~\shortcite{Shalunts2016} also performed experiments to explore the impact of MT on sentiment analysis in German, Russian and Spanish using general news articles. They surprisingly found that the performance of the sentiment analysis tool on the source and the target was comparable, which indicated that the impact of machine translation on sentiment was not obvious.
Contrary to their result, Fukuda and Jin~\shortcite{Fukuda2022} found that sentiment was significantly affected by MT tools. More specifically, positive sentences tended to be more similar in sentiment polarity before and after translation than negative and neutral sentences. Apart from the aforementioned manual or sentiment score-based evaluation of emotion translation, Saadany et al.~\shortcite{Saadany2021b} proposed a sentiment-aware measure which can be used to adjust automatic evaluation metrics like BLEU~\cite{Papineni2002} for the evaluation of MT quality of user-generated content. 

As can be seen above, most of the work does not focus on proposing a systematic human evaluation framework to assess the MT quality in terms of emotion preservation, especially not for Chinese-English translation. Our work focuses specifically on this particular use case. 

\subsection{Quality Assessment of Machine Translation}

In the MT area, there are several different automatic and human evaluation methods for assessing MT quality. Among those automatic evaluation metrics, BLEU is the most used tool for this purpose. However, BLEU has been criticised for the lack of recall and the ``explicit word-matching between translation and references''~\cite{Banerjee2005}. Other metrics like ROUGE~\cite{Lin2004} and METEOR~\cite{Banerjee2005} were proposed as an alternative to BLEU, but the resultant evaluation has been similar when compared to BLEU in terms of the n-gram matching. More recently, since the rise of BERT-like models~\cite{Devlin2018}, metrics like BERTScore~\cite{Zhang2019} have been proposed to calculate the similarity between the candidate/hypothesis and the reference translation to evaluate MT quality. 

An alternative way to measure quality is to figure out how much post-editing is needed for the candidate translation to match with the reference translation. Translation Edit Rate (TER), which is defined as ``the minimum number of edits needed to change a hypothesis so that it exactly matches one of the references, normalized by the average length of the references''~\cite{Snover2006}, is a metric that measures this error based on edit distance. 

More recently, Direct Assessment (DA)~\cite{Graham2013} of the translation output, which provides a continuous score within a certain range after the annotator sees a candidate translation and a translation hint, has been used in various ways. It can be used directly to evaluate translation quality as it is obtained from human annotators. It is also used as an input for training quality estimation models in recent Conferences of Machine Translation\footnote{https://www.statmt.org/}. Apart from DA, the MQM framework~\cite{Lommel2014} provides a more detailed evaluation methodology. It divides translation errors into six dimensions \textit{i.e.,} accuracy, fluency, design, locale convention style, terminology, and verity. Each dimension consists of several error categories like addition, mistranslation, omission or untranslated under the accuracy dimension, and more fine-grained subcategories~\cite{Lommel2018}. Each error falls into at least one of these categories and contributes to the overall rating of the translation. Error severity could be added as weights to the rating according to the seriousness of these errors. Eventually, an evaluation score can be calculated to measure the overall translation quality using the framework. The practicality, reliability, and validity of this framework~\cite{Mariana2015} have made it the choice of the translation industry and MT evaluation research. 

Nevertheless, all the above automatic methods were proposed without taking into account any elements of meaning or emotion, and human evaluation metrics were proposed for the assessment of general MT quality, which might be too generic or over-complicated for specific needs like emotion preservation. 

\section{Framework for Quality Assessment of Emotion-Loaded Texts} \label{sec3}

To evaluate the preservation of emotions, we modify the MQM framework~\cite{Lommel2014} for the assessment of MT quality of emotion-loaded microblog texts. Since our focus is on the emotion preservation, we simplify the multidimensional metrics into one dimension, \textit{i.e.,} the accuracy of translating emotions. Our error types follow the accuracy dimension of MQM, \textit{i.e.,} addition, mistranslation, omission, untranslated and source error, but we only consider errors that affect emotion. For instance, an addition error is an error in translation that adds information which does not exist in the source and the addition of this information affects the emotion in the target. Our severity levels are defined based on MQM suggestion: critical, major, and minor, which indicates how severely the source emotion is affected by the error. We define them as follows:  

\begin{itemize}
\itemsep 0mm
\item \textbf{a critical error} leads to an absolute change of emotion into a very different or even opposite emotion category;
\item \textbf{a major error} pertains to a change of emotion into one that is not very different from the original emotion or one that is somewhere between the original emotion category and another different category;
\item \textbf{a minor error} results in a slight change of emotion with uncertainties about the MT emotion label but certainties about the slight difference between the emotions of the source and the MT text.
\end{itemize}

Similar to the MQM translation quality score~\cite{Lommel2014}, we can also compute evaluation scores regarding emotion preservation by summing up all errors as per their severity level weights. Severity level weights are defined in the MQM framework and for this study, we define them as follows: 10 for critical errors, 5 for major errors and 1 for minor errors. The error rate or evaluation score of emotion translation can now be computed using Equation~\ref{eq:score}. Examples of error annotation can be seen in the Appendix. 

\begin{equation} \label{eq:score}
\begin{split}
&Error\ Rate = \frac{\sum^n_{n=1}Error_n*Weight_{s}}{Text\ Length}\\
&Weight_{s}:\ weight\ given\ to\ each\ error\\
&according\ to\ its\ severity\ level\\
&Text\ Length:\ count\ of\ all\ words\ and\\
&punctuations\ in\ the\ target\ text\\
\end{split}
\end{equation}

\section{Data and Methodology} \label{sec4}

\subsection{Data Description}

To evaluate the transfer of emotions, we need the source text to be full of emotions. The dataset for the \emph{Evaluation of Weibo Emotion Classification Technology on the Ninth China National Conference on Social Media Processing}\footnote{\url{https://smp2020ewect.github.io/}} (SMP2020-EWECT) is an ideal source for our purposes. It was annotated with six emotion categories, namely, anger, fear, joy, sadness, surprise and neutral, which was provided by the Harbin Institute of Technology and sourced from Sina Weibo~\cite{Guo2021}. 

\begin{figure}[H]
  \centering
  \includegraphics[width=0.45\textwidth]{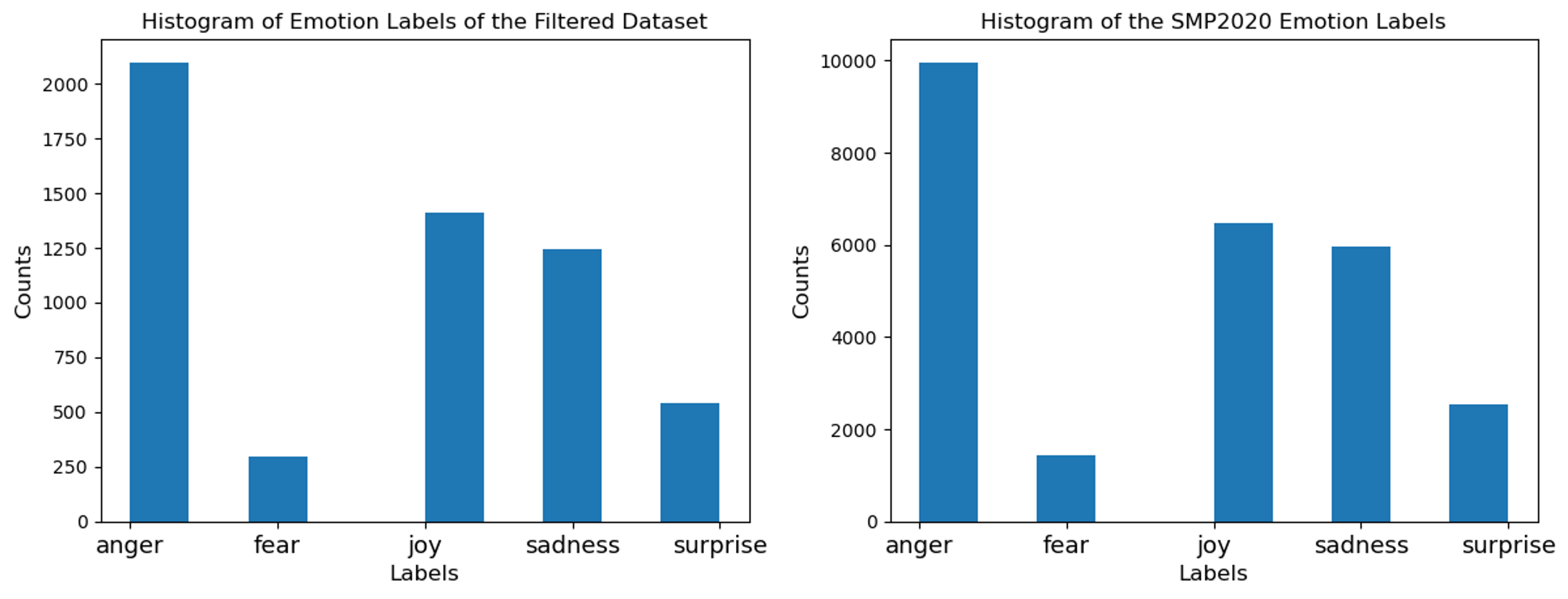}
  \caption{Distributions of Emotion Categories for the Filtered VS Original Dataset}
  \label{fig.main1}
\end{figure}

Since the dataset is as large as 34,768 entries and it includes Weibo posts with neutral emotions as well, we filter out those posts with neutral emotions and randomly sample 20 percent (about 5500 entries) for machine translation and quality assessment. The distributions of the emotion labels of our sampled dataset and the original SMP2020-EWECT dataset can be seen in Figure~\ref{fig.main1}. We can see that our sampled dataset keeps the original data distribution. We use Google Translate \footnote{Results from ``https://translate.google.co.uk/'' on the 30th of May 2022.} to translate the source text of our sampled dataset and the output is used for quality assessment. 

\subsection{Methodology}

Re-annotation of the emotions in the MT output may prove difficult in some cases due to the fact that some outputs do not make any sense for humans. For example, the MT output ``Playing this old game, I just have no friends...'' may not make much sense and it is difficult to annotate it with an emotion label. However, a bilingual annotator can easily see that the emotion of the source ``玩这个老游戏，我简直是叼到没朋友…'' which means ``Playing this old game, I'm just too good to have rivals'', is not present in the target. Therefore, we do not re-annotate the raw MT with emotion labels to check possible loss of emotions. Instead, we assess the quality of MT using the framework in Section~\ref{sec3}.

Two annotators with Chinese-English translation qualifications were recruited to annotate error types and severity levels. All translation errors coupled with severity levels that affect the transfer of original emotions were annotated in the MT output. Words or parts of the text in both source and target in relation to the translation errors were highlighted so that they can be used for error analysis. The annotators were given clear and detailed instructions about the decision process behind the annotations. We released the annotation guidelines along with the annotated dataset in our GitHub repository for inspection and reproducibility. 

Since the perception of emotion usually varies a lot among people and across time, we randomly sampled 10\% (about 550 entries) of the whole dataset for the inter-annotator agreement check and 100 entries for the intra-annotator agreement check to measure how well annotators agree with each other and themselves. The intra-annotator agreement was done by one annotator annotating the same 100 samples twice two months apart. 

\section{Result of Human Evaluation} \label{sec5}

This section shows the result of human evaluation on our Weibo dataset based on the framework and methodology proposed in previous sections. We first show the result of inter and intra-annotator agreement and then analyse the evaluation result from two aspects: 1) how many errors there are and how severe these errors are in terms of emotion category and error type; 2) what are the linguistic phenomena that are the likely cause for these errors.

\subsection{Result of the Inter and Intra-Annotator Agreement}

We use the Cohen Kappa score~\cite{Cohen1960} to calculate the inter and intra-annotator agreements. Table~\ref{tab:iaa} shows that the Kappa scores for intra-annotator agreement are very high, which means the annotator is consistent with himself/herself during annotation. Inter-annotator agreement is relatively lower, especially for the error severity. So we compared the severity levels of the two annotators and found they are more likely to disagree on whether there is a minor error (or no error). Disagreement on major/critical errors comes the second. This may be partially because different people perceive emotions differently. To further analyse the reasons, we collect some examples which annotators disagree. 

\begin{table}[H]
\normalsize
  \centering
  \resizebox{7cm}{!}{%
  \begin{tabular}{ccccc}
    \hline
&               & Error Existence  & Type  & Severity  \\ \hline  
& Inter-AA  & 0.6689        & 0.5117      & 0.3691       \\ 
& Intra-AA  & 0.8991        & 0.8990      & 0.7634      \\    \hline
  \end{tabular}%
  }
  \caption{Cohens Kappa for Inter and Intra-Annotator Agreement (AA) for Error Existence, type and severity.}
  \label{tab:iaa}
  \end{table}

One of the main causes is the disagreement on the change of the subject of emotion. For example, the MT output of the source ``吓死宝宝了'' meaning ``Scared me to death'', is ``Scared the baby to death''. One annotator annotates it as a minor error, while the other as a major error. In this example, the subject of emotion should be ``me'' rather than a third party, ``the baby'', which might result in the reduction of the strong emotion and the transformation of the emotion from ``fear'' into somewhere between ``fear'' and ``anger''. Annotators are likely to disagree on the severity level of this case. 

\begin{figure*}[t]
  \centering
  \includegraphics[width=0.7\textwidth]{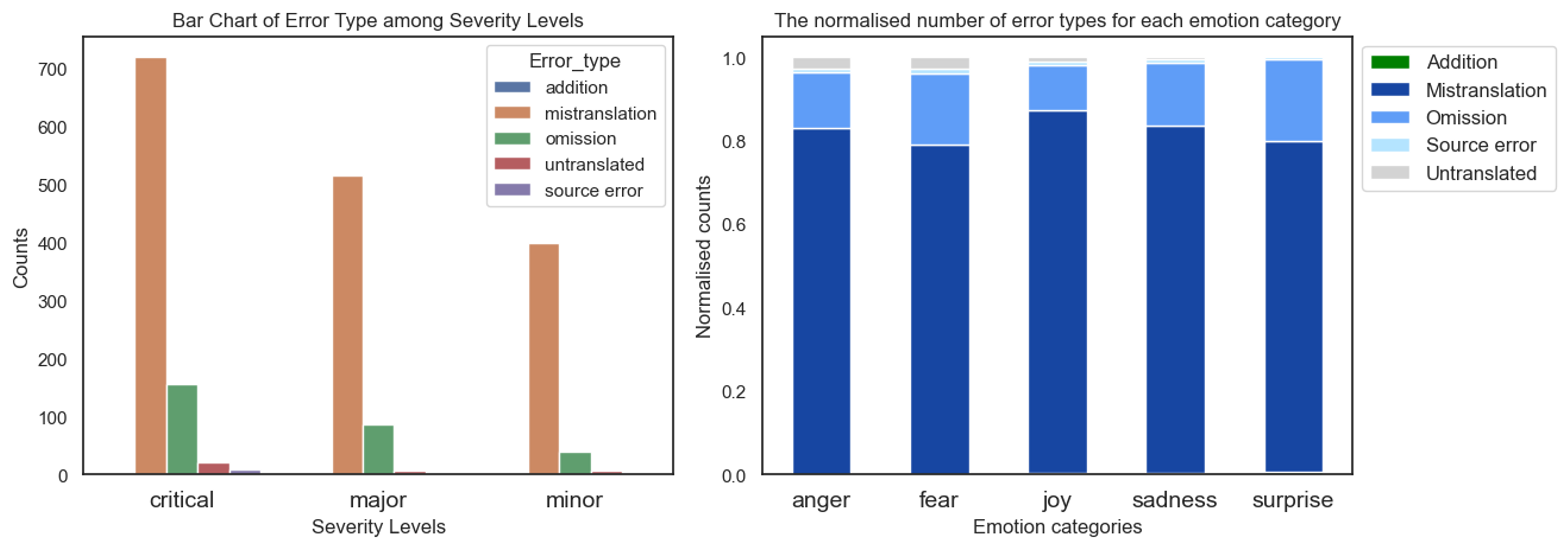}
  \caption{Error Types against Severity Levels and Emotion Categories where the first chart (left) shows the error types among severity levels and the second shows normalised counts for error types among emotion categories.}
  \label{fig.main3}
\end{figure*}

Emotion conflicts caused by mistranslation is another problem which annotators disagree. For instance, the source emotion of this post ``我容易嘛我~黑眼圈,青春痘,眉毛,皱纹~全在这两天爆出来了'' is sadness, which means ``Life is so hard on me. Dark circles, pimples, eyebrows, wrinkles all had an explosive growth in the past two days'', but the MT output ``I'm easy. I~ Dark circles, pimples, eyebrows, wrinkles~ have all exploded in the past two days'' may contain both joy and sadness, two conflicting emotions. This causes the disagreement on the severity level, as one annotator annotates it as a critical error, while the other as a major error. 

The complete change of meaning in the target but with the similar emotion as the source is another major cause. For example, the emotion of the MT output ``His mother got a leg and caught a cold again, mad at me'' might be anger or sadness, which is similar to the emotion of the source ``他娘了个腿的，又感冒了，气死我了'', but the target meaning is completely different from the source ``F**k your mother, Cold again! I'm so pissed off''. One annotator annotates it as a critical error, while the other as a major error. 

\subsection{Error Statistics}

After annotating each entry of the dataset, we collect all error entries and display error statistics in the following figures to see 1) how many examples are incorrectly translated; 2) which type of error is most common; 3) which emotion category is less likely to be mistranslated; and 4) which error type is more critical. 

\begin{figure}[H]
  \centering
  \includegraphics[width=0.9\columnwidth]{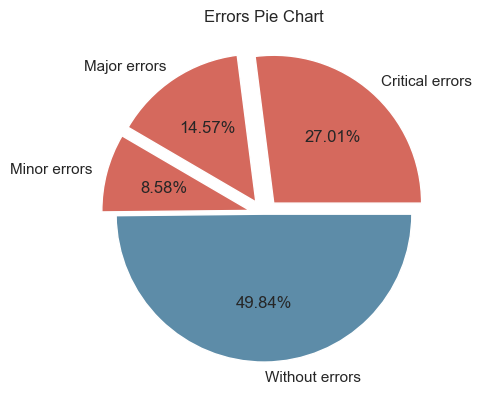}
  \caption{Error Severity in overall MT output}
  \label{fig.main2}
\end{figure}

From Figure~\ref{fig.main2}, we know the MT quality of these texts is not acceptable as about 50\% of the entries have errors in preserving emotions and 41.58\% have major or critical errors. 

\begin{figure*}[t]
  \centering
  \includegraphics[width=0.7\textwidth]{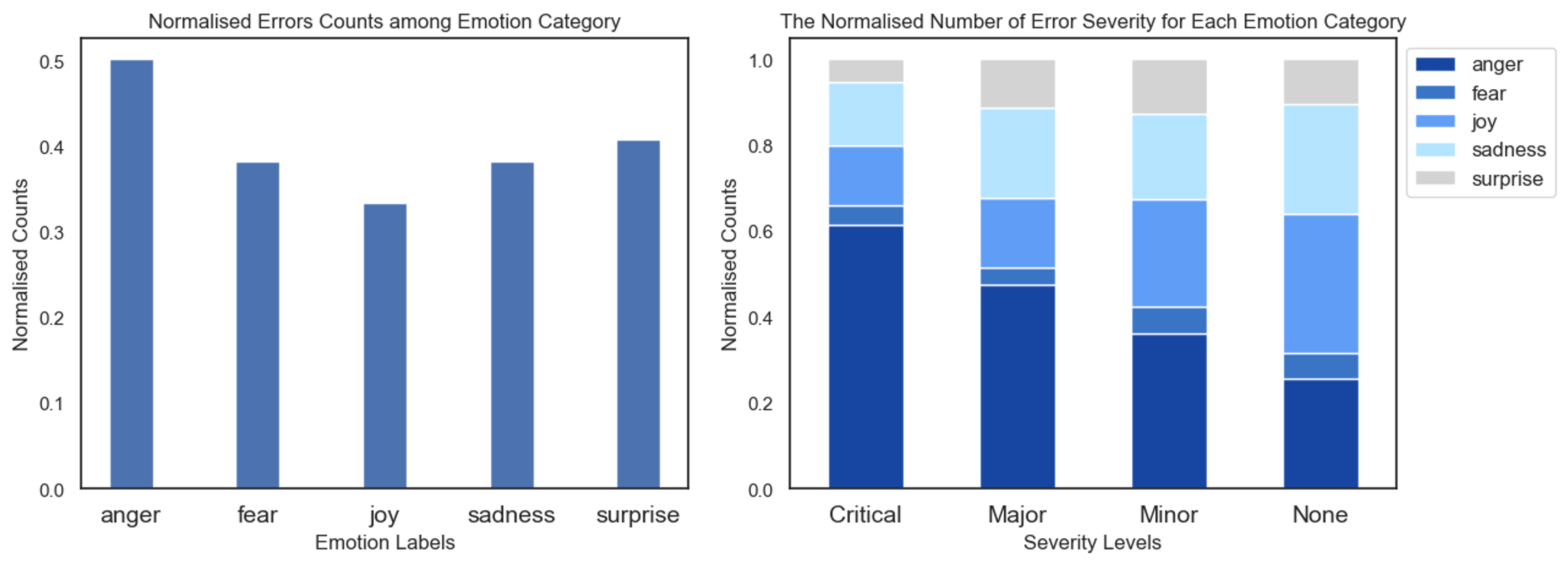}
  \caption{Errors among Emotion Categories where the first chart (left) shows the error proportion among emotion categories whereas the second chart shows normalised counts of severity levels among emotion categories.}
  \label{fig.main4}
\end{figure*}

Among these error severity levels, mistranslation is the most common error type followed by omission according to the left chart in Figure~\ref{fig.main3}. In the right bar chart of Figure~\ref{fig.main3}, we normalise the number of error types of each emotion category against the total number of errors. We can see the pattern is very similar for all emotion categories, which suggests mistranslation is the most common error type and omission comes the second. 

In the left bar chart of Figure~\ref{fig.main4}, the number of errors is divided by the number of instances in each emotion category to show the proportion of errors in that emotion category. We see that `joy' accounts for the least errors despite it having the second largest number of total entries, which means that those social media texts with the emotion of `joy' are more likely to be translated correctly by Google Translate, compared with other emotion categories. This can be further proved by the right chart of Figure~\ref{fig.main4}, where normalised counts of severity levels are plotted for each emotion category. We can see from critical errors to no error, as the severity level decreases, the number of `joy' increases. This suggests errors in the `joy' category are more likely to be minor. For those entries without errors, `joy' takes the largest percentage among all emotion categories. This result corresponds with the study by Fukuda and Jin~\shortcite{Fukuda2022}, which indicated that positive sentences are less likely to be affected by MT compared with negative and neutral sentences. 

\begin{figure}[H]
  \centering
  \includegraphics[width=0.9\columnwidth]{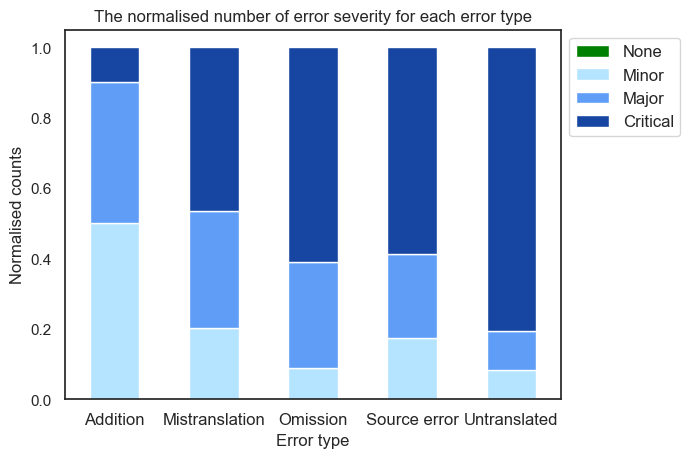}
  \caption{Normalised Error Severity in Error Types}
  \label{fig.main5}
\end{figure}

In Figure~\ref{fig.main5}, we normalise the number of error severity for each error type against the total number of errors. We can see that for all error types, critical errors take the largest percentage except for addition. In the addition category, minor errors are much more than critical errors, which means addition errors are less likely to have severe impact on emotions. That is maybe because the original emotion would not be changed a lot if we just add some extra words in the target text. For the untranslated category, critical errors are far more than other types. This suggests that untranslated errors affect the transfer of emotion quite severely. 

\subsection{Analysis of Error Causes}

In this section, we investigate linguistic phenomena that are responsible for the translation errors in the MT output based on annotation described in Section~\ref{sec4}. We first discuss errors caused by emotion carrying words and then by other linguistic phenomena. 

\subsubsection{Emotion Carrying Words}

To find out the most common cause of these translation errors, we collect all the words and sentences identified during annotation as corresponding to an error and then find out where the error occurs. We count the frequency of these words and sentences, and calculate the percentage of the words in total erroneous entries as shown in Table~\ref{tab:main3} and Table~\ref{tab:main4}. 

\begin{table}[H]
\tiny
  \centering
  \resizebox{7.5cm}{!}{%
  \begin{tabular}{ccccc}
    \hline
   & Source        & Frequency      & Human Translation            & Word Percentage  \\ \hline  
   & 尼玛           & 50           & (f**k) your mother             &  2.19\% \\ 
   & 居然           & 42           & actually                       &  4.37\% \\ 
   & 竟然           & 22           & surprisingly                   &  2.96\%  \\ 
   & 特么           & 20           & what's the f**k                &  1.86\%   \\ 
   & TMD           & 14           & WTF                            &  0.58\%   \\
   & TM            & 14           & WTF                            &  1.29\%   \\
   & 还是           & 12           & still                          &  5.59\%  \\
   & 真是醉了        & 12           & really speechless              &  0.45\%  \\
   & 日了狗了        & 10           & f**ked up                      &  0.39\%  \\
   & 折腾           & 10           & mess around                    &  0.64\%  \\
   & 草泥马         & 10           & f**k your mother               &  0.71\% \\ \hline
  \end{tabular}%
  }
  \caption{Most Frequent Words in Erroneous Examples}
  \label{tab:main3}
  \end{table}

We can see from ``Human Translation''\footnote{Human translations here and in the rest of the paper are provided by a professional translator.} column in Table~\ref{tab:main3} that almost all the frequent words are emotion carrying words. Some of them, including the most frequent word ``尼玛'', are emotional slang created by homophone character substitution~\cite{Chu2017}. Others such as ``居然'', ``竟然'' are emotional adverbs used to show strong feelings. Many of these emotion carrying words (top five) take a large percentage among all erroneous entries. For example, ``尼玛'' appears in 2.19\% of the erroneous entries in emotion translation. 

\begin{table}[H]
\tiny
  \centering
  \resizebox{7cm}{!}{%
  \begin{tabular}{ccccc}
    \hline
   & Source          & Frequency      & Human Translation                  \\ \hline  
   & 我也是醉了        & 12            & I'm really speechless       \\ 
   & 吓死宝宝了        & 8             & scared me to death           \\ 
   & 我tm快炸了        & 4            & I'm f**king exploding         \\ 
   & 不要不要的        & 4             & super/very                 \\ 
   & 服了自己了        & 4             & disappointed to myself        \\ \hline
  \end{tabular}%
  }
  \caption{Most Frequent Phrases in Erroneous Examples}
  \label{tab:main4}
  \end{table}

Table~\ref{tab:main4} shows the most frequent 5 phrases among those erroneous examples. We can see that these phrases also contain slang or adverbs that convey strong emotions. From both tables, we observe that emotion carrying words pose a strong challenge to translation. 

\subsubsection{Other Linguistic Phenomena}

Other linguistic phenomena like polysemous words, abbreviation, negation, subject/object issues, subjunctive mood and punctuation problems \textit{etc.}, also play a role in causing these errors in emotion translation.

\paragraph{Polysemous Words}

Polysemous words especially those having several different meanings can be easily mistranslated, which might result in the change of the original emotion. In the following example, the character ``疼'' in the source literally means ``hurt'', but in the Chinese culture, it can represent an emotion called ``heart-aching love'' which refers to the love that children get from their doting parents or lovers get from their partners~\cite{Sundararajan2015}. MT clearly mistranslates the source emotion.

\noindent Source Text (ST): 介个女人说会\textbf{疼}我一辈子

\noindent Machine Translation (MT): Tell a woman that she will \textbf{hurt} me for the rest of my life

\noindent Human Translation (HT): This woman said she will \textbf{love} me for the rest of her life.

\paragraph{Abbreviation}

Internet slang in Chinese can be created by abbreviation, which shortens a longer expression into a word/phrase. In the source of the following example, ``活久见'' literally meaning ``live long see'' is an abbreviation of ``\textbf{活}的时间\textbf{久}什么事都可能\textbf{见}到'', which is often used to imply surprise. Mistranslation of this abbreviation by MT leads to the misunderstanding and change of the source emotion. 

\noindent ST: \textbf{活久见}，我还是比较适合高冷。就一个人喜欢我萌。晚安

\noindent MT: \textbf{See you for a long time}, I am still more suitable for high cold. The only one who likes me is cute. Good night

\noindent HT: \textbf{If you live long enough, you can see anything unexpected}. I am more suitable for being cool. Only one person sees me as cute. Good night.

\paragraph{Negation}

Mistranslation of negation is a known problem for MT affecting both the emotion preservation and the understanding of a text. In the following example, the source character ``好'' means ``very'' not the common meaning of ``good'' and ``不'' is the negative word, but in the MT result, only ``好'' is kept as ``good'' not the correct meaning of ``very'' and the negation is omitted. 

\noindent ST: 心情\textbf{好不}爽

\noindent MT: I'm in a \textbf{good} mood

\noindent HT: I'm in a \textbf{very bad} mood.

\paragraph{Subject/Object Issues}

Since Chinese is not a subject prominent language~\cite{Tan1991}, omission of subject is a quite common phenomenon in Chinese especially in informal texts. The omission of the subject in the source causes the swap of the subject and object in MT and results in a change of the emotion subject. This further affects the emotion of the MT as it becomes closer to fear rather than anger. 

\noindent ST: 拉我一下能死吗

\noindent MT: Can I die if I pull

\noindent HT: Will you die if you pull me up?

\paragraph{Subjunctive Mood}

Chinese does not have syntactic markers for counterfactual conditionals as the subjunctive mood in English~\cite{Feng2006}. The source text expresses the wish to run the first place, but machine translation does not render it into the English subjunctive mood, affecting the transfer of the original anger emotion. 

\noindent ST: 再跑不到第一把在我前面的都删了

\noindent MT: I can't run the first one. I deleted the one in front of me.

\noindent HT: If I didn't run the first place, I would delete all those who run ahead of me.

\paragraph{Punctuation Problems}

Nonstandard use of punctuation in Chinese microblogs is another challenge posed to emotion translation. Here, the following source text is separated by exclamation marks, which shows strong emotions. But in the MT output, each separated character is regarded as an independent sentence. Such mistranslations change the original emotion, as the character ``好'' meaning ``very'' is translated as ``good''. 

\noindent ST: 我！好！饿！！！！！

\noindent MT: I! it is good! hungry! ! ! ! !

\noindent HT: I AM SO HUNGRY!!!!!

\vskip 0.2in

The following example shows problems caused by the lack of punctuation. Since there is no space between Chinese characters, it is difficult for MT systems to tokenise the sentence. The lack of punctuation in some entries in the dataset seems to be highly correlated with the quite frequent omission of some emotion loaded parts in the text.

\noindent ST: 到底什么时候去考试啊老是忽悠我再拖下去没心情去考试

\noindent MT: When are you going to take the test

\noindent HT: When are we going to take the exam? Always fooling me. I would be in a bad mood if it postponed again.

\paragraph{Hallucination}

Hallucination~\cite{Lee2018} is a common problem for neural machine translation, but it is rarely seen in this dataset. We only see the following example of hallucination, which might probably be caused by continuous repetition of some characters since the MT result keeps changing as we edit the repetitive characters. Hallucination is definitely a problem for the preservation of the source emotion.

\noindent ST: 次奥次奥次奥次奥次奥次奥次奥次奥次奥次奥次奥次奥次奥次奥次奥次奥次奥次奥次奥次奥真特么是醉了

\noindent MT: 200022000

\noindent HT: WTF WTF WTF WTF WTF WTF WTF WTF WTF WTF WTF WTF WTF WTF WTF WTF WTF WTF WTF WTF I'm f**king speechless.

\section{Conclusion and Future Work} \label{sec6}

Our work investigates the performance of MT engines on the translation of emotion-loaded texts. We propose a new framework for evaluating MT quality in terms of emotion preservation developed in line with the MQM evaluation framework. We perform a manual evaluation of the MT output and present a detailed error analysis. We observe which type of errors is the most common and which emotion category is more likely to be correctly translated by MT. Our detailed analyses describe which linguistic factors such as emotion carrying words, subject omission and so on, cause these errors in translating microblog texts loaded with emotions. Furthermore, the annotated bilingual dataset can be used for training quality estimators to automatically assess the translation quality while preserving emotions. In future, we aim to extend this dataset with reference translations and use it to train computational models for estimating the translation quality of emotion-loaded texts. We plan to conduct further research and perform more analyses to improve the proposed framework. 

\bibliography{eamt23}
\bibliographystyle{eamt23}
\end{CJK}
\end{document}